\relax
\documentclass[letterpaper]{article} 
\usepackage{aaai21}  
\usepackage{times}  
\usepackage{helvet} 
\usepackage{courier}  
\usepackage[hyphens]{url}  
\usepackage{verbatimbox}
\usepackage{graphicx} 
\usepackage{natbib}  
\usepackage{caption} 
\usepackage{tikz}
\newcommand*{\circled}[1]{\lower.7ex\hbox{\tikz\draw (0pt, 0pt)%
    circle (.5em) node {\makebox[1em][c]{\small #1}};}}
\usepackage{dialogue}

\nocopyright

\title{Task and Situation Structures for Service Agent Planning}
\author {
    Hao Yang\textsuperscript{\rm 1}
    Tavan Eftekhar\textsuperscript{\rm 1}
    Chad Esselink\textsuperscript{\rm 1}
    Yan Ding\textsuperscript{\rm 2}
    Shiqi Zhang\textsuperscript{\rm 2}
}
\affiliations {
    
    \textsuperscript{\rm 1} Ford Motor Company, USA \\
    \textsuperscript{\rm 2} State University of New York - Binghamton, USA \\
    hyang1@ford.com
}

\begin{document}

\maketitle

\begin{abstract}
%
%
Everyday tasks are characterized by their varieties and variations, and frequently are not clearly specified to service agents. This paper presents a comprehensive approach to enable a service agent to deal with everyday tasks in open, uncontrolled environments. We introduce a generic structure for representing tasks, and another structure for representing situations. Based on the two newly introduced structures, we present a methodology of situation handling that avoids hard-coding domain rules while improving the scalability of real-world task planning systems. 
\end{abstract}

\section{Introduction}
\label{sec:introduction}

\noindent Classical task planning is to develop a series of actions that transform the world from an initial state to a state that includes the goals of the task \cite{ghallab2016automated}. It also assumes that a complete task plan is generated before executing the plan \cite{fikes1971strips, aeronautiques1998pddl}. However, in the real world, people often do not know the exact details of every future state at the planning time. Also, the world state can be changed by exogenous events at execution time.  So, people plan at different levels of abstractions, leaving long-range plan details out at the planning time. Hierarchical planning was introduced to reflect this intuitive planning practice \cite{tate1977generating, sacerdoti1974planning, yang1990formalizing, nau2003shop2}. In Hierarchical Task Network (HTN) planning, refinement rules, called \emph{methods}, break down a task into sub-tasks. These planning techniques are predominantly rule-driven.  

\citeauthor{scholnick1993planning} (\citeyear{scholnick1993planning}) studied several use cases of task planning. 
They used “Tower of Hanoi”~\cite{havur2013case} to exemplify “problem-solving tasks” and used “errand planning task” to exemplify “everyday tasks.” They compared and contrasted similarities and differences, and found that the nature of errand planning is quite different from solving Tower of Hanoi. For instance, although errand planning is not as computationally demanding, its planning environment is much more variable. Errand plans rarely play out in a perfectly predictable environment. Goals for errands are often not hard-and-fast or even not well defined. Many more factors influence the execution of a task plan. The rule-based approach that works well with problem-solving tasks finds many difficulties in everyday tasks. \citeauthor{ghallab2014actor} (\citeyear{ghallab2014actor}) have also contrasted the differences between planning in the highly abstracted model and executing in a complex real-world environment. The factotum robot in \cite{ghallab2014actor} is a use case similar to the use case of “errand planning.” The intricacies in planning in such use cases post significant challenges to the traditional state search approaches. 

We study a \emph{service agent} that interacts with human clients for assigned tasks. In this scenario,  the agent faces seemingly unlimited kinds of tasks plus their variations, and task plans. The agent may need to repair a plan if the plan fails or need to respond to a “situation.”  

Many researchers have addressed the plan repair problem. For example, ASPEN \cite{rabideau1999iterative, chien2000using} has a plan repair mechanism developed for Mars rovers. The plan repair unit keeps monitoring conflicts and applies repair methods when conflicts are detected. ASPEN has a total of ten repair methods. The Refinement Acting Engine (RAE) \cite{ghallab2016automated, patra2019acting} takes an acting-oriented approach that makes online planning a part of an action. It keeps searching for an acting method from a comprehensive refinement method library based on the states of the world, qualified conditions, and the task. Goal Driven Anatomy (GDA) \cite{ghallab2014actor} was proposed that has a four-phase process of anomaly detection and goal modification/generation process in response to environmental changes or imperfect predictions. It continuously monitors any anomaly and generates new goals that the planner will use to revise the existing plan or create new plans. All these approaches are relying on hand-crafted rules, which are primarily domain dependant.   

We define a \emph{Situation} as “an unexpected event or a demand that the agent needs to respond to" in this paper. There are various \emph{Situations} that could happen. They are unpredictable, and it is hard to enumerate all of them. Those rare \emph{Situations} are often referred to as “corner cases" or “edge cases". Take robotaxi as an example. Assuming that a vehicle picks up a customer and sends the customer from location $A$ to location $B$. Many Situations can happen during the trip. At the pick-up location, the vehicle agent and the customer may not find each other, or the vehicle could not access the pre-arranged spot. During the trip, the customer may complain about the smell or spill in the car, or the customer needs to divert for an urgent errand. We are particularly interested in the use case of “errand planning” in this research. 
Here are two hypothetical Situation examples:
\vspace{3mm}

\hangindent 0mm
\hangafter=0
\noindent
\textbf{The story of “Window Leak"}

\footnotesize{

\begin{quote}
    
\direct{It starts raining.
Passenger Annie saw water seep into the cabin. The window is not fully closed.
}

\begin{dialogue}
\speak{Annie} The water is getting in.

\direct{The vehicle agent checks the window, one of them is open. 

The agent sends a command to the control unit to close the window. However, the window is not closed.

The agent realizes that the window is in a malfunction. 

The agent recalls a case that the window that could not get closed because the window glass was blocked by a twig.}
\speak{vehicle} Is there something that blocked the window glass?
\speak{Annie} Yes, looks like it is jammed

\direct{The agent rolls down the window a little bit and the rider cleaned a foreign object that jammed the window. 
The vehicle agent rolls up the window again and the window is closed this time. }

\end{dialogue}
\end{quote}

}

\normalsize{}

\noindent
\textbf{The story of “pharmacy”}

\begin{quote}
    
\footnotesize{
\direct{Passenger Joe went on a business trip. He rides in a vehicle towards the hotel. He passed by a pharmacy and realized that he can pick up a prescription there.}
}
\begin{dialogue}

\speak{Joe} Could you stop by that pharmacy?

\direct{The vehicle requests the Map Agent to find a pharmacy that is on the way to the hotel. The vehicle shows the map location of a pharmacy on the screen in the vehicle. }

\speak{vehicle}Do you want to go to this pharmacy?.
\speak{Joe} No, I’d like to go to the one we just passed. \direct{Joe only wants to go the pharmacy he just passed by.}

\direct{The Map Agent presents more nearby pharmacies on the map on the screen.}

\speak{vehicle} How about these?

\direct{Joe points to the one he wants to go on the touch screen.}
\speak{vehicle} Will you come back and continue your trip?
\speak{Joe} Yes.
\speak{vehicle} How long should I wait?
\speak{Joe} Maybe 10 to 15 minutes.
\speak{vehicle} I will wait for you at the front door of the store in 10 minutes.

\direct {The vehicle turns around and drives to the pharmacy.

The vehicle offboards Joe at the pharmacy.

10 minutes later, the vehicle will be back to resume the trip to the hotel.
}

\end{dialogue}
\end{quote}

\normalsize{}

\emph{Situations}  like these that could be solved relatively easily by a human driver become challenges to an AI agent. Not only \emph{Situations} are numerous, but also the difference in the context of \emph{Situations} compounds variations. The solution space is impossible to be exhaustively defined. It is infeasible to hard-code all the rules to solve them. 

This paper illustrates a comprehensive design and practice that aims to provide the following solutions:
\begin{enumerate}
    \item It creates text-based, generic structures and syntax for Tasks and Situations of all variations. 
    \item It embeds domain knowledge in executed cases to avoid the necessity of hard-coded or static domain rules.
    \item It offers a design that tolerates imperfections in the model (goals, conditions, states, actions, effects) and data. It uses context as additional attributes into guiding the search for solutions.

\end{enumerate}

In the following sections, we first introduce a generic \emph{Task} structure we created. We then present a generic \emph{Situation} structure and \emph{situation handling} process and methodology. After that, we illustrate how the two example \emph{Situation} cases could be implemented with the situation handling process while leveraging our task and situation structures.

\section{Task Structure}

Our connotation of a \emph{Task} is more than its goal conditions. We include the whole history of how this \emph{Task} is accomplished into the \emph{Task} with a motivation of reusing the \emph{Task}. There are many kinds of \emph{Tasks} for a service agent, big or small (big \emph{Tasks} can be divided into small \emph{Tasks}). We created a generic \emph{Task} structure that suits all \emph{Tasks}. 
 
 \emph{First}, we define tasks and sub-tasks recursively. They share the same \emph{Task} structure. \emph{Second}, the \emph{Task} structure encapsulates all parameter details in the \emph{Task} and the task plan. This approach follows the theory of case-based reasoning (CBR) and case-based planning \cite{aha1996case, hammond2012case}.  A \emph{Task} is a record of history, an episode of a story. The \emph{Task} structure is implemented with a syntax that is intuitive to engineers (using json format\footnote{“json is a lightweight data-interchange format." For details, please check: https://www.json.org/json-en.html}). \emph{Third}, it introduces context as an attribute of a \emph{Task}. Context plays an essential role in planning \cite{scholnick1993planning,  leake2014context}, especially for a service agent. Instead of pattern-matching-based solution search, which fits the Tower of Hanoi use case, it is more beneficial to use similarity-based retrieval for a often imprecisely defined problem space like the errand planning use case. Context is an additional attribute to influence the similarity calculation.

\subsection{Task Structure}

The \emph{Task} structure is illustrated in Table~\ref{table}. It follows the concept of classic planning with a few attributes that may be worth explaining:

\begin{table}[h]
\caption{Main attributes of the \emph{Task} Structure}
\vspace{-5mm}
\begin{center}
\begin{tabular}{r@{\quad}ll}
\hline
\multicolumn{1}{l}{\rule{0pt}{12pt}Attribute}& 
\multicolumn{2}{l}{Explanation}\\[2pt]
\hline\rule{0pt}{12pt}
\fontsize{9pt}{9pt}
\textbf{Task\_name}:    & string name of a \emph{Task} class  & \\
\fontsize{9pt}{9pt}
\textbf{Parent\_task}:  & null if no parent \\
\fontsize{9pt}{9pt}
\textbf{Sub-tasks}: &   a list of sub-tasks, empty if leaf & \\
\fontsize{9pt}{9pt}
\textbf{Action}: & the action of the \emph{Task} & \\
\fontsize{9pt}{9pt}
\textbf{Specs}:  & detail specs for the action & \\
\fontsize{9pt}{9pt}
\textbf{Conditions}:  & preconditions for this \emph{Task} & \\[2pt]
\fontsize{9pt}{9pt}
\textbf{Effects}:  & effects after the \emph{Task} is performed & \\[2pt]
\fontsize{9pt}{9pt}
\textbf{Context}:  & a list of contexts of this \emph{Task}& \\[2pt]
\fontsize{9pt}{9pt}
\textbf{Goals}:  & goals to be verified & \\[2pt]
\fontsize{9pt}{9pt}
\textbf{Est Time}:  & estimated execution time & \\[2pt]
\hline
\end{tabular}
\end{center}
\label{table}
\end{table}

\textbf{Sut\_tasks}: A list of sub-tasks. Each sub-task takes the same structure of a \emph{Task}. 

\textbf{Action} is the abstract form of a \emph{Task}. An Action has the action (verb), and a few parameters of this action (like a predicate). For example, Action: “\texttt{\fontsize{9pt}{9pt}\selectfont Robot-r move block-A from location-1 to location-2}”. The Action “move” takes three arguments: object-to-be-moved, start-location, and end-location, plus the actor, recorded under a schema attribute. The \emph{Task} of “move”, however, has more information such as \textbf{Conditions}, \textbf{Goals}, etc. 

\textbf{Specs} contains the details of parameters that are used in the action. For example, if \texttt{\fontsize{9pt}{9pt}\selectfont location-1}, a label, is used as the origin location in a trip domain application, then the location object (with details such as GPS coordinates, type of location, etc.) is included in the \textbf{Specs}. 

\textbf{Conditions} may have different types: “hard” conditions have to be satisfied before the \emph{Task} could be performed; “fail-skip” conditions are conditions that, if failed, the \emph{Task} could be skipped. Other types of conditions could be defined such that when such a condition fails during \emph{Task} execution, it generates a context.

\textbf{Context} contains any context information relevant to this \emph{Task}. For example, if the Task is to go to the San Francisco airport, going “by bus" or going by “driving a personal car" is context information of this \emph{Task}. 

This schema is implemented in json. We have \emph{serialize} and \emph{deserialize} functions in Python that transform data to objects or objects to data when needed. So, we can encode objects in data in a naturally understandable form.  

We have built a system for illustration and discussion of \emph{Task} structure, assuming a robo driver serving a customer, called Virtual Service Agent (VSA). 

\subsection{Execution of \emph{Tasks}}

In VSA, task planning is an integral part of the task execution process. Before a \emph{Task} is executed, it develops its sub-tasks. When a \emph{Task} develops its sub-tasks, VSA does not apply a method as HTN planning does. Instead, VSA takes a variational approach: it copies an existing \emph{Task} with sub-tasks instantiated and replaces the Specs and Contexts from the new Task. The existing \emph{Task} could be from a template or a previously executed \emph{Task}.  In the current implementation, When a \emph{Task} is created, it takes the initial status of \texttt{\fontsize{9pt}{9pt}\selectfont unplanned}. After the planning stage, when a \emph{Task} develops its sub-tasks, the \emph{Task} changes its status to \texttt{\fontsize{9pt}{9pt}\selectfont planned}.  A Task may be of other status: \texttt{\fontsize{9pt}{9pt}\selectfont executing}, \texttt{\fontsize{9pt}{9pt}\selectfont finished}, \texttt{\fontsize{9pt}{9pt}\selectfont failed},  \texttt{\fontsize{9pt}{9pt}\selectfont aborted}, etc.).
In the next stage, if there are sub-tasks, each sub-task is iterated and its \texttt{\fontsize{9pt}{9pt}\selectfont execution} function is recursively called the same way as the parent \emph{Task}. If there are no sub-tasks, the action is executed, which usually is sending the \emph{Task} to another agent (the actor) for execution. 

During execution, When a \emph{Task} develops its sub-tasks, the “specs" in the sub-tasks will be mapped from the parent \emph{Task}’s “specs". The bindings of parameters from the parent to children are through a mapping attribute in the \emph{Task} structure (not shown in Table \ref{table}). The following is an example of a \emph{mapping}:

\vspace{1mm}
\hspace{-5mm}
\begin{verbbox}[\fontsize{7.5pt}{7.5pt}\selectfont]
{ 
  "spec.origin”: "parent.specs.origin”,
  "specs.destination”: "parent.specs.destination”
}
\end{verbbox}
\hspace{1mm}
\fbox{
\theverbbox
}
\vspace{0.5mm}

It means that the \texttt{\fontsize{9pt}{9pt}\selectfont origin} in the \emph{specs} of this \emph{Task} is assigned the same as the \texttt{\fontsize{9pt}{9pt}\selectfont origin} of the parent \emph{Task} specs. The \texttt{\fontsize{9pt}{9pt}\selectfont destination} in the specs of this \emph{Task} is also assigned the same as the \texttt{\fontsize{9pt}{9pt}\selectfont destination} of the parent \emph{Task} specs.

If there is an exception detected during the plan execution, the exception is handled based on the error message. Some of the exceptions will be considered as \emph{Situations} and \emph{Situations} will be handled by the agent. If the \emph{Task} could not be executed, (for example, if the conditions are not satisfied, and a \emph{Situation} could not be handled successfully), the \emph{Task} status will be changed to \texttt{\fontsize{9pt}{9pt}\selectfont failed}. The \texttt{\fontsize{9pt}{9pt}\selectfont failed} status will propagate to its parent \emph{Tasks}, all the way up. Every executed \emph{Task} is saved to the \emph{Task} case library to retain a rich \emph{Task} repository.

This \emph{Task} structure naturally supports replay and simulation. Each executed \emph{Task} is archived in the case library, and we can replay it in the future. On the other hand, we can simulate a \emph{Task} plan similar to replay. It is imperative to have a simulation system in a case-based planning system \cite{hammond2012case}. Once an old plan is modified, it is not guaranteed to succeed. A robust simulation will detect failures so that flaws in the modified plan can be repaired. The simulation acts as a  validation. The details of validation will be explained later in Section~\ref{sec:situationhandling}: Situation Handling Process. 

\subsection{Implementation}

\begin{figure*}[t]
  \centering
  \includegraphics[width=\textwidth]{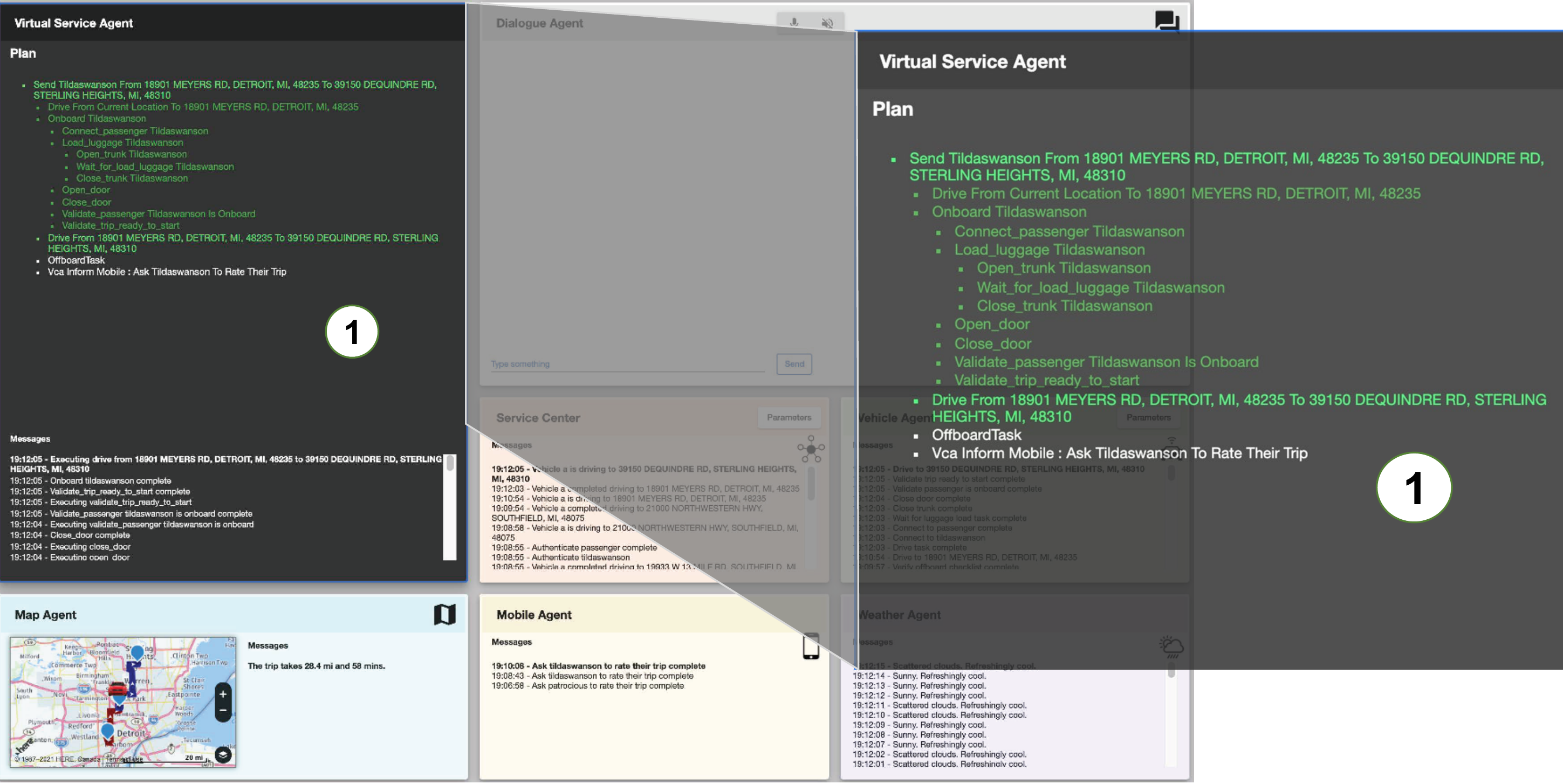}
  \caption{VSA User Interface}
  \label{fig:vca}
\end{figure*}

 Figure~\ref{fig:vca} shows the implementation of our VSA (Virtual Service Agent) System. In the graphical user interface, each window represents an agent. Within an agent window, there is an action panel at the top and a message panel at the bottom. The top left window is the VSA panel, where we can monitor the Task plan as it is executing. The lower left window is the Map Agent that simulates the vehicle driving through a trip. Among other agents, we have a Dialogue Agent that communicates using natural language with the rider, a Weather Agent that retrieves live weather information, a Mobile Agent that emulates the communication to the rider through a mobile device, a Vehicle Agent that controls the vehicle mechanics and sensors, the Service Center that is the dispatch system that sends trip tasks to the vehicle. 

Figure~\ref{fig:vca}\circled{1} is an example of the \emph{Task} hierarchy of a \emph{Trip Task}. Each line prints the action of the \emph{Task}. We implement it to resemble a \emph{Trip Task} handled by a vehicle sending a customer, \texttt{\fontsize{9pt}{9pt}\selectfont Tildaswanson}, a fictitious name, from location Meyers Rd to location Dequindre Rd. The \emph{Trip Task} is received from a trip assignment platform (the Service Center). A \emph{Trip Task} has four top-level sub-tasks: A Drive \emph{Task} that drives from where the vehicle is to the pickup location Meyers Rd; at Meyers Rd, the Agent performs the Onboard \emph{Task}; it then performs a Drive Task that drives from Meyers Rd to Dequindre Rd; after arriving Dequindre Rd, it performs the Offboard \emph{Task}. The sub-task Onboard \emph{Task}, for example, has its sub-tasks: connect-passenger, load-luggage, etc. The load-luggage \emph{Task} is further developed into sub-tasks: open-trunk, wait-for-load-luggage, close-trunk. Certainly, whether having the load-luggage \emph{Task} depends on if the customer has luggage that needs to be put in the trunk. This information is captured in the context information of the parent \emph{Task}. Instead of using rules like “if has-luggage then $\dots$" in the refinement method as you would expect in an HTN planning system, VSA uses \emph{Task} attributes, including contexts, as indices to search for a previous similar \emph{Task} as a template to develop the sub-tasks.

\section{Situation Structure}

In an open or semi-open world, a service agent faces multifarious \emph{Situations}. Exhaustively define \emph{Situations} in \emph{ad hoc} fashion does not scale.   
Here, we present a generic \emph{Situation} structure that is capable of describing all \emph{Situations} and \emph{situation handling} without domain specific data types and code. \emph{Situation} types and situation handling knowledge are not hard-coded, but recorded in text format (json strings) and is in data. 

In a nutshell, a \emph{Situation} will be handled using a \textbf{Remedy} to repair the plan. However, a \emph{Situation} (class) may be handled with different \textbf{Remedies}, depending on the context. A \emph{Situation} has a \textbf{Context} attribute that is used to differentiate variations of a specific \emph{Situation} class. It follows the case-based reasoning methodology such that situation handling cases can be reused. There is also a \textbf{Logics} field in the \emph{Situation} structure. It is intended to embed problem-solving knowledge in text-based data so that the knowledge is not hard-coded.

\begin{table}[h]
\caption{Situation Structure}
\vspace{-3mm}
\begin{center}
\begin{tabular}{r@{\quad}ll}
\hline
\multicolumn{1}{l}{\rule{0pt}{12pt}
Attribute}& \multicolumn{2}{l}{Explanation}\\[2pt]
\hline\rule{0pt}{12pt}
\fontsize{9pt}{9pt}
\textbf{Name}:    & name of this \emph{Situation} & \\
\fontsize{9pt}{9pt}
\textbf{Time}:  & time this \emph{Situation} occurred & \\
\fontsize{9pt}{9pt}
\textbf{Task}: & Task during which the \emph{Situation} is logged & \\
\fontsize{9pt}{9pt}
\textbf{Context}: & contexts while this \emph{Situation} happened \\
\fontsize{9pt}{9pt}
\textbf{Remedy}:  & a list of remedy actions to take & \\
\fontsize{9pt}{9pt}
\textbf{Logics}:  & how to set the Context and the Remedy& \\[2pt]
\fontsize{9pt}{9pt}
\textbf{Goals}:  & new goals the repaired plan should satisfy & \\[2pt]
\hline
\end{tabular}
\end{center}
\label{situation}
\end{table}

Table~\ref{situation} is the \emph{Situation} structure. Here are the fields:

\begin{itemize}
    \item 
\textbf{Name}, the \emph{Situation} name is considered as the class name of similar \emph{Situations}. For example, if the window of the car is broken, the Situation name could be “car window broken”.

\item The \textbf{Time} field is self-explanatory. 

\item \textbf{Task} is the \emph{Task} that was executing when the \emph{Situation} was logged or the \emph{Task}  during which the \emph{Situation} is handled. There might be multiple Tasks the agent is performing when the \emph{Situation} is logged, in the case of parallel  \emph{Task}  processing. In this case, the most relevant  \emph{Task}  is used. For example, if the vehicle is driving and the vehicle is playing music, a \texttt{\fontsize{9pt}{9pt}\selectfont car\_window\_broken} \emph{Situation} will reference the \texttt{\fontsize{9pt}{9pt}\selectfont driving}  \emph{Task} . 

\item \textbf{Context} is a list of context attributes under which this \emph{Situation} occurs. For example, “it is raining" could be a context attribute of a “\texttt{\fontsize{9pt}{9pt}\selectfont car\_window\_broken}” \emph{Situation}. The severity of the window glass broken could be another context attribute of this \emph{Situation}.

\item \textbf{Goals} are a list of new goals that need to be satisfied if the \emph{Situation} is handled successfully.
\end{itemize}

The above information is received when a \emph{Situation} is detected or received (from another agent), we call it the \emph{Situation} header. 

\begin{itemize}

    \item
\textbf{Logics} is used to help determine the contexts that are most relevant to this \emph{Situation}. The context knowledge may be used for situation handling.  For example, in the \texttt{\fontsize{9pt}{9pt}\selectfont car-window-broken} \emph{Situation}, the \textbf{Logics} will request a sensor agent to find which window is broken, the severity of the damage, a weather agent the current weather condition. In the implementation, \textbf{Logics} are a list of functions that feed into the \textbf{Context}. The following is an example of \textbf{Logics}. It is in the form of a (python) dictionary:

\vspace{1mm}
\begin{verbbox}[\fontsize{8pt}{8pt}\selectfont]
"logics": {
    "window_broken": "vda.checking_window",
    "weather": "weather.current_weather",
    "wetness”: "chat.wetness”
}
\end{verbbox}
\fbox{
\hspace{-3mm} \theverbbox
\quad
}
\vspace{1mm}

In this example, the keys are attributes that will appear in the context. The values are the functions. The first function is a function of the “vda" agent, which has sensors to tell if a window is malfunction or broken. The second function is a function of a weather agent. It returns the current weather condition. The third function initiates a Chat conversation, getting how much of the concern of the wetness in the cabin from the passenger in the vehicle. The return of the function is a free-formed text string. These attributes are added to the Context information of this \emph{Situation}. The functions could be more sophisticated, and examples of them are outside of the scope of this paper.

\item  \textbf{Remedy} is a list of \emph{remedy action}s used to alter the task plan so that the \emph{Situation} is resolved.   A \emph{remedy action} (or in short: \emph{remedy}) is simply adding/deleting/modifying a Task. 

\end{itemize}

\begin{table}[h]
\caption{\emph{Remedy action} Structure}
\vspace{-5mm}
\begin{center}
\begin{tabular}{r@{\quad}ll}
\hline
\multicolumn{1}{l}{\rule{0pt}{12pt}
Attribute}& \multicolumn{2}{l}{Explanation}\\[2pt]
\hline\rule{0pt}{12pt}
\fontsize{9pt}{9pt}
\textbf{Operation}:    & add/delete/modify & \\
\fontsize{9pt}{9pt}
\textbf{Reference}:  & a list defines references of attributes & \\
\fontsize{9pt}{9pt}
\textbf{Mapping}: & functions that fills spec of with\_task & \\
\fontsize{9pt}{9pt}
\textbf{With\_task}: & new task that will be added or modified \\
\hline
\end{tabular}
\end{center}
\label{remedy}
\end{table}

Table~\ref{remedy} shows details of a remedy action structure.
In the \emph{remedy action} structure:

\begin{itemize}
\item Operation: an example operation will be something like: “\texttt{\fontsize{9pt}{9pt}\selectfont add after the drive\_task}"; or “\texttt{\fontsize{9pt}{9pt}\selectfont modify this\_task}". It contains both an operation (\texttt{\fontsize{9pt}{9pt}\selectfont add/modify/delete}) and the target information (“\texttt{\fontsize{9pt}{9pt}\selectfont after the drive\_task}" / “\texttt{\fontsize{9pt}{9pt}\selectfont after this\_task}", etc.). We adopt this natural syntax. It can be easily parsed with a set of vocabulary.  
\item References: A list of reference definitions that connect attributes with objects in the program.

Through “references”, the keys in the \emph{mapping} are referenced to the actual object in the program. In the following example:

\begin{verbbox}[\fontsize{8pt}{8pt}\selectfont]
"references": {
    "drive_task": "executing task",  
    "context": "situation context"  
}
\end{verbbox}

\fbox{{
\hspace{-3mm}
\theverbbox
\quad}
}

“\texttt{\fontsize{9pt}{9pt}\selectfont drive\_task}" that is used in the \emph{mapping} is referenced to the “\texttt{\fontsize{9pt}{9pt}\selectfont executing task}” (the “Task” in Table~\ref{situation}); “\texttt{\fontsize{9pt}{9pt}\selectfont context}" that is used in the \emph{mapping} is referenced to the \textbf{Context} in the \emph{Situation} (Table~\ref{situation}).

\item Mapping: how the \texttt{\fontsize{9pt}{9pt}\selectfont Specs} of the new Task (the “\texttt{\fontsize{9pt}{9pt}\selectfont with\_task}” in Table~\ref{remedy} is to be set. 

The following is an example of the \emph{mapping}: 

\begin{verbbox}[\fontsize{7.5pt}{7.5pt}\selectfont]
"mapping”: {
  "specs.origin": "drive_task.specs.origin",
  "specs.dest": "context.current_location",
  "specs.actor": "drive_task.actor",
  "action.origin": "drive_task.specs.origin",
  "action.dest": "context.current_location",
  "estimated_time": "drive_task.actual_duration"
}
\end{verbbox}
\fbox{
\hspace{-3mm}\theverbbox

}

In each mapping item, the left side of “:” (key) is the target of the parameter, the right side of “:” (value) is the source of the parameter. Please notice that “\texttt{\fontsize{9pt}{9pt}\selectfont drive\_task}” and “\texttt{\fontsize{9pt}{9pt}\selectfont context}” in the source parameters are defined in the “\texttt{\fontsize{9pt}{9pt}\selectfont references}” described just above.

\item With\_task: the new \emph{Task} that is to be added into the task plan.
\end{itemize}

\section{Situation Handling Process}
\label{sec:situationhandling}

When a \emph{Situation} is detected, the agent will retrieve the \textbf{Logics} of this \emph{Situation}  (class) from the \emph{Situation} library. The \textbf{Logics} functions are invoked, and the return values will populate additional \textbf{Context} information of the \emph{Situation}. The \emph{Situation} with its Context is then pushed to a Situation Queue.

When the agent executes a \emph{Task}, it also keeps checking if there is any \emph{Situation} in the Situation Queue. If there is a \emph{Situation} in the Queue, the agent will attempt to handle the \emph{Situation}. The agent will first use the \emph{Situation} name and \textbf{Context} to retrieve any prior \emph{Situation} in the Situation library that matches best with the \emph{Situation}. If a similar \emph{Situation} is found, the \textbf{Remedy} of the old \emph{Situation} will be used to repair the plan of the new \emph{Situation}.

\begin{figure}[h]
\centering
\includegraphics[width=0.9\columnwidth]{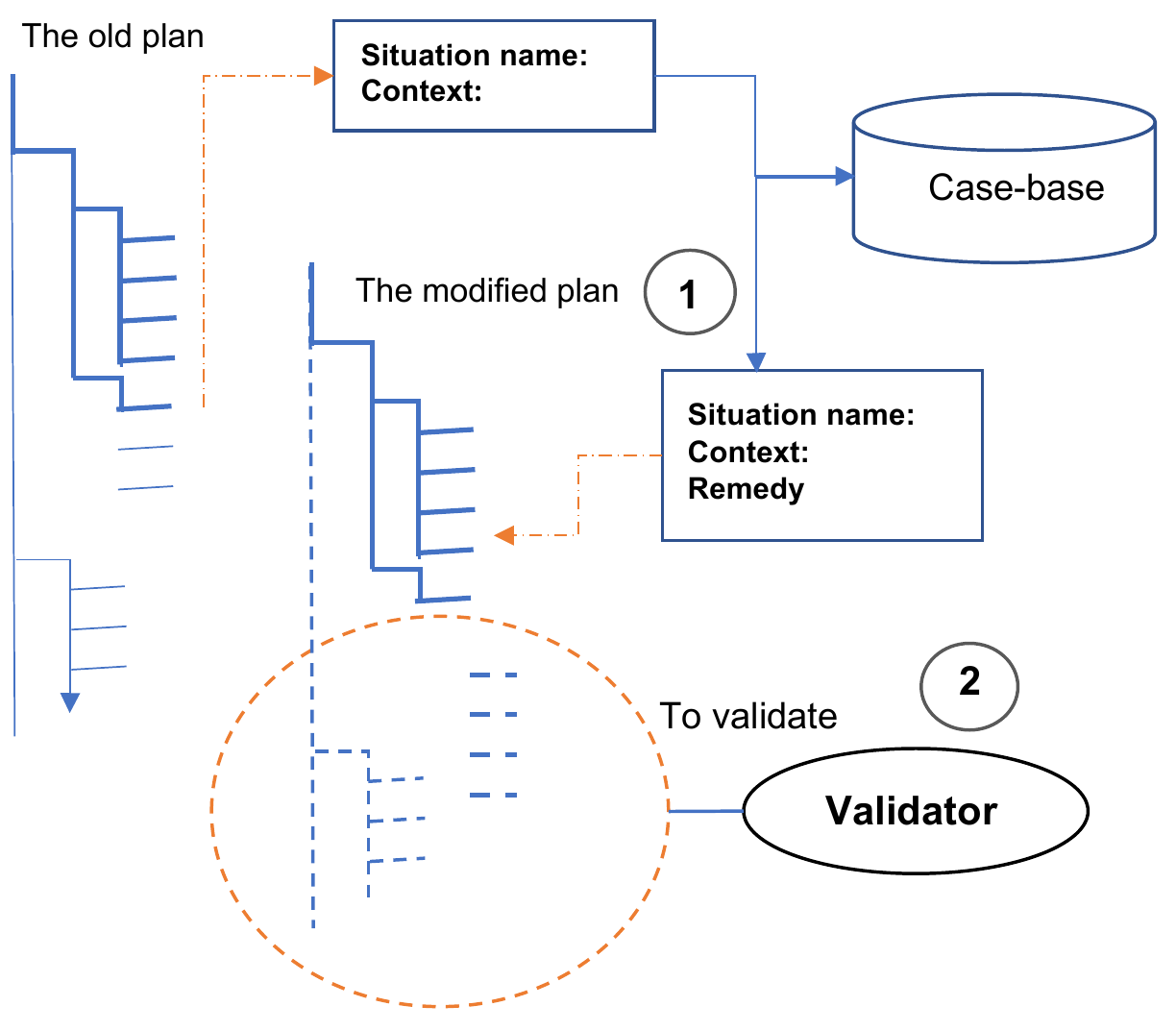} 
\caption{An example of plan modification}
\vspace{-1mm}
\label{fig:modification}
\end{figure}

Once the \textbf{Remedy} is applied, the modified plan (Figure~\ref{fig:modification}\circled{1}) will be validated using the Validator (Figure~\ref{fig:modification}\circled{2}). In Figure~\ref{fig:modification}, the solid lines in the modified plan are the Tasks that have been executed. The dashed lines are the \emph{Tasks} that have not been executed. The validation is to validate the unexecuted \emph{Tasks}. The validation is through a simulation process. It starts with the current State. The agent simulates each \emph{Task} by checking the conditions first, and then it executes the \emph{Task} in the simulation mode, and it applies the effects of the \emph{Task} to the States, and then it checks if the goals of the \emph{Task} are met.

If the goals are met to the end, the modified plan is validated. Otherwise, there are two options: one is to find another similar \emph{Situation} case in the \emph{Situation} library to repair the plan and validate the repaired plan again. Another is to call in human assistance as in the process described below.

What if there is a \emph{Situation} that VSA does not know beforehand? What if there is no prior \emph{Situation} that is similar enough (to pass a similarity threshold) to the new \emph{Situation}? In this case, human intervention is inevitable. However, what we want is that a new \emph{Situation} class can be easily introduced, and a new Remedy can be easily constructed. We also want that the new situation handling case can be reused in the future. Next, we will discuss this process.

\begin{figure}[h]
\centering
\includegraphics[width=1.0\columnwidth]{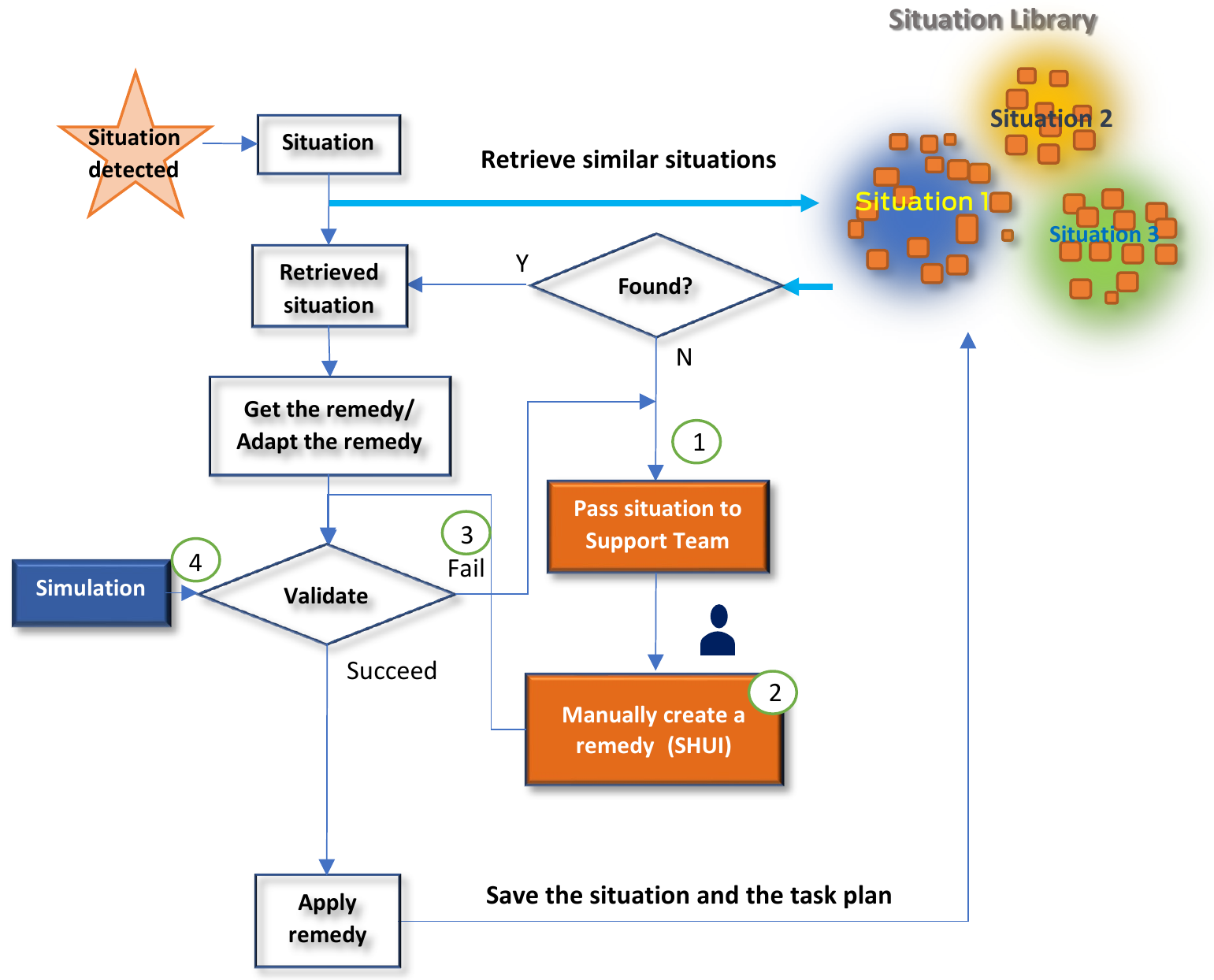} 
\caption{Situation handling life cycle process}
\vspace{-1mm}
\label{fig:lifecycle}
\end{figure}

\begin{figure}[hpt]
\centering
\includegraphics[width=1.0\columnwidth]{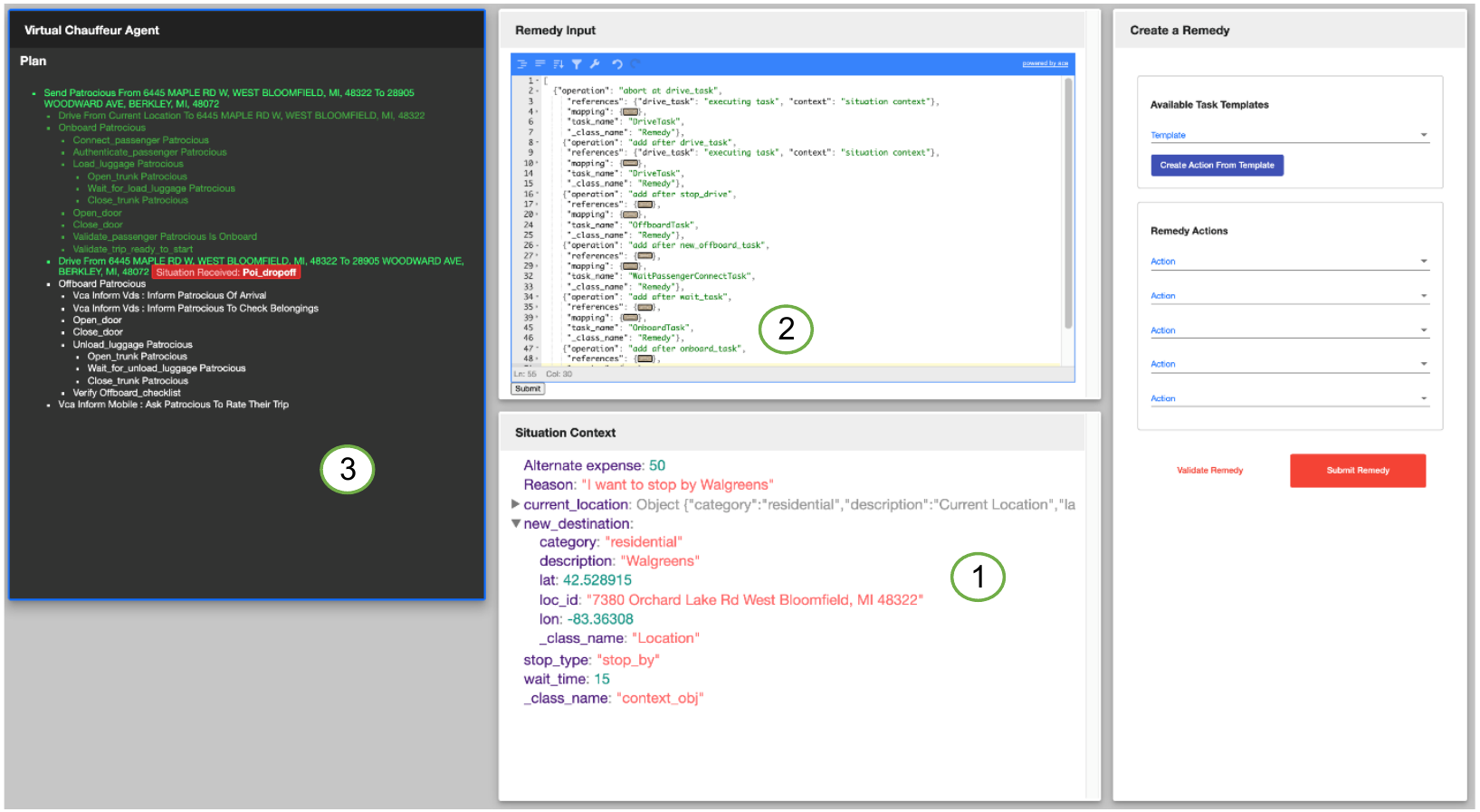}
\caption{Situation Handling UI}\label{fig:shui}
\end{figure}

Figure~\ref{fig:lifecycle} depicts this process. Figure~\ref{fig:lifecycle}\circled{1} is when the remote customer assistant center is informed. A customer assistant will be able to quickly see the current status of the \emph{Situation} (“what is the \emph{Situation}?", “when did the \emph{Situation} happen?", “the contexts of the \emph{Situation}?", and “the Specs of the \emph{Task}?"). The customer assistant can directly talk to the customer to find out the additional context that helps him/her to resolve the \emph{Situation}. The customer assistant will do all these through the system called Situation Handling UI (SHUI) as shown in Figure~\ref{fig:shui}.  SHUI is a comprehensive user interface that allows trained customer support experts (support specialists) to craft new Remedies in the integrated system. SHUI abstracts away specific complexities of the underlying codebase so that the support specialist can focus on problem-solving. 

Figure~\ref{fig:shui} is an example of the SHUI interface. The left panel (Figure~\ref{fig:shui}\circled{3}) displays the real-time \emph{Task} execution that is identical to the VSA (Figure~\ref{fig:vca}). The lower-middle panel displays the situation context (Figure~\ref{fig:shui}\circled{1}). The support specialist can see exactly what is happening and what has happened at the vehicle remotely. The right panels are pallets that the support specialist pick, draw and drop Tasks, “remedy actions". The upper-middle panel shows the revised \textbf{Remedy} and the “submit” button will send the revised \textbf{Remedy} to VSA to repair the plan(Figure~\ref{fig:shui}\circled{2}). 

\section{Illustrative Examples}

Here we show how “Window Leak” and “Pharmacy”, two \emph{Situation} examples introduced in Section~\ref{sec:introduction}, are handled in VSA to illustrate the proposed situation handling methodology. 

\vspace{3mm}
\noindent
\textbf{Example 1: The story of ``Window Leak"}
\vspace{2mm}

\normalsize{Here is what happens:}

Passenger Kelly gets in the car and the vehicle starts the journey. It starts to rain. However, water seeps through the window, and water drops onto Kelly. 
“It is raining and it is wet here", Kelly claimed. 

A wet-in-cabin $Situation$ is generated. The Logics of the $Situation$ is:

\vspace{1mm}
\begin{verbbox}[\fontsize{8pt}{8pt}\selectfont]
"logics": {
    "window_broken": "vda.checking_window",
    "weather": "weather.current_weather",
    "wetness”: "chat.wetness”
}

\end{verbbox}
\fbox{
\theverbbox
}
\vspace{1mm}

“window-broken” – It calls the Vehicle Agent to check if any window is broken;

“weather” – from the Weather Agent, it returns current weather condition;

“wetness” – it initiates a dialog with the passenger to obtain the following information: wherein the cabin is wet (seat? floor? on the person?).

The above information feeds into the Context of the \emph{Situation}.
VSA finds a similar \emph{Situation} from its \emph{Situation} library that has the following remedy:

\vspace{1mm}
\begin{verbbox}[\fontsize{8pt}{8pt}\selectfont]
”add close-window task”
”add confirm-problem-solved task” 
\end{verbbox}
\fbox{
\theverbbox
}
\vspace{1mm}

The “close-window” task is sent to the vehicle, and the vehicle sends a “close-window” command.  

The “confirm-problem-solved” Task will trigger a dialog using the Dialog Agent. It returns the confirmation and related response in the form of context. 

Unfortunately, assuming, the confirmation is negative. The water is still pulling in. A new “window-fail-to-close” \emph{Situation} is created because the “close-window" \emph{Task} was confirmed as failed by the rider in the “confirm-problem-solved" \emph{Task}.

The Logics under “window-fail-to-close” \emph{Situation} is: 

\vspace{1mm}
\begin{verbbox}[\fontsize{8pt}{8pt}\selectfont]
"logics": {
  ”close_window": "vda.close_wdw_status",
  ”window_malfunc": "vda.wdw_malfunc_detect",
  ”window_broken”: ”vda.broken_wdw_detect”
}
\end{verbbox}
\fbox{
\theverbbox
}
\vspace{1mm}

In the above Logics, the “close\_window” context is already filled from the previous situation handling process. Therefore, the context is carried over. 

Assuming we have the following contexts (in addition to all other contexts we have had) after applying the Logics:

\vspace{1mm}
\begin{verbbox}[\fontsize{8pt}{8pt}\selectfont]
"context": {
    "close_window": true,
    "window_malfunc": false,
    "window_broken”: false
}
\end{verbbox}
\fbox{
\theverbbox
}
\vspace{1mm}

A similar \emph{Situation} was found that has Remedy:

\begin{verbbox}[\fontsize{8pt}{8pt}\selectfont]
”add confirm-passenger task: window-is-jammed”  
\end{verbbox}
\fbox{
\theverbbox
}

The answer populates the Context. Assume that the Context is: \texttt{\fontsize{9pt}{9pt}\selectfont "window-is-jammed”: true}. 

A new similar \emph{Situation} “window-is-jammed” is found and the remedy is:

\vspace{1mm}
\begin{verbbox}[\fontsize{8pt}{8pt}\selectfont]
”add open-window task”
”add request passenger task: remove foreign
     object"
”add close-window task” 
”add confirm-problem-solved task” 
\end{verbbox}
\fbox{
\hspace{-2mm}
\theverbbox
}
\vspace{1mm}

Assuming the final confirmation is positive, and the \emph{Situation} is resolved. The newly logged \emph{Situation} and the history will be saved to the \emph{Situation} library and Task library. 
In case the final confirmation is negative, and the Agent could not find a relevant \emph{Situation}. In that case, VSA may send the \emph{Situation} to SHUI, and human intervention will be called to resolve the \emph{Situation}.

\vspace{3mm}
\noindent
\textbf{Example 2: The story of “Pharmacy"}
\vspace{2mm}

Here is what happens in VSA for our next example:

The Dialogue Agent posts a “\texttt{\fontsize{9pt}{9pt}\selectfont POI\_dropoff}" \emph{Situation} (POI - point-of-interest) on the Situation Queue.

When VSA receives the “\texttt{\fontsize{9pt}{9pt}\selectfont POI\_dropoff}" \emph{Situation} on the Situation Queue, it attempts to handle the \emph{Situation}. 

The Situation Header looks like this:

\vspace{1mm}
\fbox{
\footnotesize{}
    \parbox{0.8\linewidth}{
        \textbf{Situation Name}: \texttt{\fontsize{9pt}{9pt}\selectfont POI\_dropoff}
        
        \textbf{Task}: \texttt{\fontsize{9pt}{9pt}\selectfont Drive\_task}
        
        \textbf{Context}: \{
        
\hspace{4mm}\textbf{current\_location}: {location …},

\hspace{4mm}\textbf{stop\_location}: {location…},

\hspace{4mm}\textbf{stop\_type}: \texttt{\fontsize{9pt}{9pt}\selectfont "stop\_by"},

\hspace{4mm}\textbf{wait\_time}: \texttt{\fontsize{9pt}{9pt}\selectfont 15}

\}
}
}
\vspace{1mm}

\normalsize{
The situation handling finds a previous  “\texttt{\fontsize{9pt}{9pt}\selectfont POI\_dropoff}" \emph{Situation} in the database. The Context of the retrieved old \emph{Situation} has “\textbf{stop\_type}" of “\texttt{\fontsize{9pt}{9pt}\selectfont final destination}", which means the passenger would choose the "poi-stop" as her final destination, she would not continue her original journey. The final destination of the trip was changed to the “\textbf{stop\_location}" of the \emph{Situation}, defined in the Context. The retrieved \emph{Situation} has three \emph{remedy actions} in the \textbf{Remedy}:  

\vspace{1mm}
\begin{verbbox}[\fontsize{7.5pt}{7.5pt}\selectfont]
 {"abort at drive_task"...},
 {"add after current_drive_task"...},
 {"modify at next_offboard_task"...}
\end{verbbox}
\fbox{
\theverbbox
}

\normalsize{}

\begin{quote}
\begin{enumerate}
\vspace{-1mm}
\item aborts the current \texttt{\fontsize{9pt}{9pt}\selectfont drive\_task}; 
\item adds a \texttt{\fontsize{9pt}{9pt}\selectfont drive} task to the new \textbf{stop-location}; 
\item modifies the \texttt{\fontsize{9pt}{9pt}\selectfont offboard\_task} so that the offboard location is the new \textbf{stop-location}. 
\end{enumerate}
\end{quote}

\begin{figure*}[t]
    \centering
    \includegraphics[width=\textwidth]{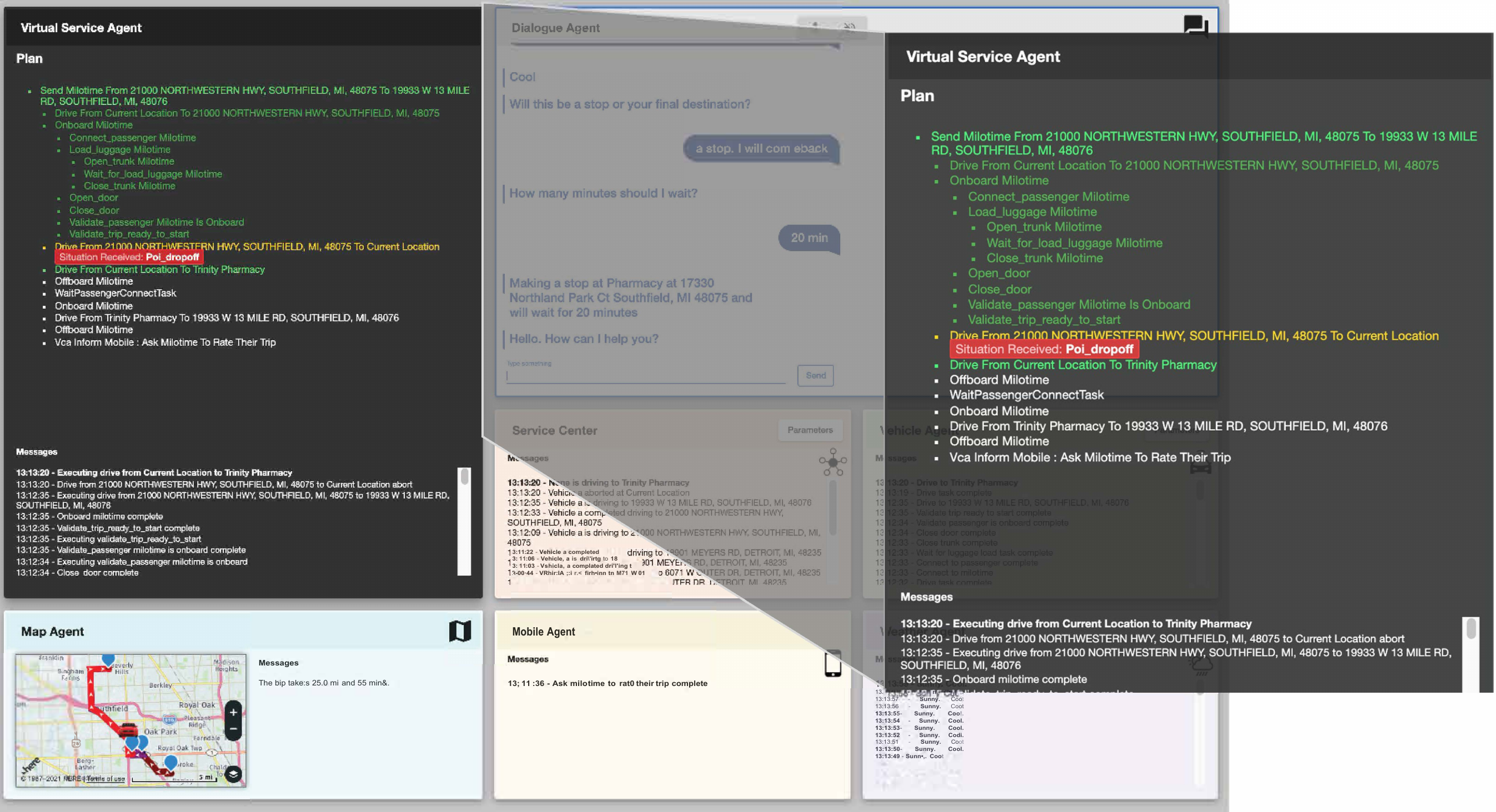} 
    \caption{The repaired plan}
    \label{fig:poi}
\end{figure*}

The final destination of the trip was changed to the “\textbf{stop\_location}", defined in the Context.

The new “\texttt{\fontsize{9pt}{9pt}\selectfont  POI\_dropoff}" \emph{Situation}, however, is different such that the passenger will continue his journey to his original destination. This is defined in the goal of the \emph{Situation}. }

When the \textbf{Remedy} of the retrieved Situation was adapted to the new “\texttt{\fontsize{9pt}{9pt}\selectfont  POI\_dropoff}" Situation, it encounters an exception in the validation (Figure~\ref{fig:modification}\circled{3}). Because the goals of the new situation are different. One of the new goals is that the final destination is the same as the original destination, instead of the stop-location. This exception is  captured in the validation and VSA will send the Situation to SHUI. A new \textbf{Remedy} is created manually and is sent back (Figure~\ref{fig:modification}\circled{2}, Figure~\ref{fig:shui}\circled{2}) to VSA. The new \textbf{Remedy} has six \emph{remedy actions}:

\vspace{1mm}
\begin{verbbox}[\fontsize{7.5pt}{7.5pt}\selectfont]
 {"abort at drive_task"...},
 {"add after current_drive_task"...},
 {"add after stop_drive"...},
 {"add after new_offboard_task"...},
 {"add after wait_task"...},
 {"add after onboard_task"...}
\end{verbbox}
\fbox{
\theverbbox
}

\vspace{2mm}
\vspace{-1mm}
\begin{quote}
\begin{enumerate}

\item aborts the current drive\_task; 
\item adds a drive task (stop\_drive) to drive to the "stop-location"; 
\item adds an offboard task at the "stop-location";
\item adds a wait task after the offboard task;
\item adds an onboard task after the wait\_task;
\item adds a drive task after the onboard task that drives to the final destination.
\end{enumerate}
\end{quote}

Applying this \textbf{Remedy}, the new plan passes validation. Figure~\ref{fig:poi} shows the repaired plan. You may zoom in 
Figure~\ref{fig:poi} to see the details. In the upper left VSA panel, the red shows the Situation, the yellow shows the Task that the Situation was handled; the bright green is the executing Task (corresponds to \emph{remedy action} 2); the next four white Tasks correspond to \emph{remedy action} 3-6.   After the Situation is handled, the new Situation with the revised \textbf{Remedy} is saved to the Situation library so that next time, similar Situations will be handled without human intervention.

\section{Conclusions}

A service agent deals with problems similar to everyday tasks. In such a domain, tasks have more variations; the environment is more dynamic; and goals are often negotiable or even not well defined. In such a domain, situations are multifarious as well and often fortuitous. Furthermore, context plays an important role in how a plan is developed and how a situation should be handled. In such a domain, the bottleneck lies in knowledge acquisition and representation, not in computation efficiency. 

This paper presents a comprehensive approach that addresses the nature of such a problem domain. Here we highlight some of the unique features and contributions covered in this paper: 
\begin{enumerate}
    \item Proposed a single, unified task representation structure that fits all tasks and a single situation representation structure that fits all situations for an open world domain. Explicitly called out that task, subtask, and actions are analogous, depending only on an abstraction perspective; 
    \item Proposed structure and syntax for \emph{Tasks} and \emph{Situations} so that domain rules can be completely embedded in task and situation cases; 
   \item Proposed to use context as an important attribute to reflect variations in \emph{Tasks} and \emph{Situations}.

\end{enumerate}

\bibliography{ref}

\end{document}